\documentclass[sigconf]{acmart}

\usepackage{multirow}
\usepackage[ruled,vlined,linesnumbered]{algorithm2e}
\SetEndCharOfAlgoLine{}
\usepackage{enumitem}
\usepackage{etoolbox}

\usepackage{subfigure}
\usepackage{colortbl}
\usepackage{marvosym}

\AtBeginDocument{%
  \providecommand\BibTeX{{%
    \normalfont B\kern-0.5em{\scshape i\kern-0.25em b}\kern-0.8em\TeX}}}

\copyrightyear{2024}
\acmYear{2024}
\setcopyright{rightsretained}
\acmConference[CIKM '24]{Proceedings of the 33rd ACM International Conference on Information and Knowledge Management}{October 21--25, 2024}{Boise, ID, USA}
\acmBooktitle{Proceedings of the 33rd ACM International Conference on Information and Knowledge Management (CIKM '24), October 21--25, 2024, Boise, ID, USA}
\acmDOI{10.1145/3627673.3679543}
\acmISBN{979-8-4007-0436-9/24/10}

\makeatletter
\patchcmd{\maketitle}{\@copyrightpermission}{
   \begin{minipage}{0.3\columnwidth}
     \href{https://creativecommons.org/licenses/by/4.0/}{\includegraphics[width=0.90\textwidth]{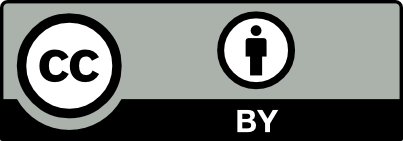}}
   \end{minipage}\hfill
   \begin{minipage}{0.7\columnwidth}
     \href{https://creativecommons.org/licenses/by/4.0/}{This work is licensed under a Creative Commons Attribution International 4.0 License.}
   \end{minipage}
   \vspace{5pt}
}{}{}
\makeatother

\begin{document}

\author{Kun Peng}
\affiliation{%
  \institution{Institute of Information Engineering, Chinese Academy of Sciences,}
  \city{Beijing}
  \country{China}}
\affiliation{%
  \institution{School of Cyber Security,
 University of Chinese Academy of Sciences,}
  \city{Beijing}
  \country{China}}
\email{pengkun@iie.ac.cn}

\author{Lei Jiang}
\affiliation{%
  \institution{Institute of Information Engineering,
 Chinese Academy of Sciences,}
  \city{Beijing}
  \country{China}}
\email{jianglei@iie.ac.cn}
\authornote{Corresponding author}

\author{Qian Li}
\affiliation{%
  \institution{Beijing University of Posts and Telecommunications,}
  \city{Beijing}
  \country{China}
}
\email{li.qian@bupt.edu.cn}

\author{Haoran Li}
\affiliation{%
  \institution{Hong Kong University of Science and Technology,}
  \city{HongKong}
  \country{China}
  }
\email{hlibt@connect.ust.hk}

\author{Xiaoyan Yu}
\affiliation{%
  \institution{Beijing Institute of Technology,}
  \city{Beijing}
  \country{China}
  }
\email{xiaoyan.yu@bit.edu.cn}

\author{Li Sun}
\affiliation{%
  \institution{North China Electric Power University,}
  \city{Beijing}
  \country{China}
}
\email{ccesunli@ncepu.edu.cn}

\author{Shuo Sun}
\affiliation{%
  \institution{Beihang University,}
  \city{Beijing}
  \country{China}
}
\email{sun.shuo@buaa.edu.cn}

\author{Yanxian Bi}
\affiliation{%
  \institution{China Academy of Electronic and Information Technology,}
  \city{Beijing}
  \country{China}
}
\email{biyanxian@cetc.com.cn}

\author{Hao Peng}
\affiliation{%
  \institution{Beihang University,}
  \city{Beijing}
  \country{China}
}
\email{penghao@buaa.edu.cn}

\theoremstyle{definition}
\newtheorem{define}{Definition}[]

\title{Table-Filling via Mean Teacher for Cross-domain Aspect Sentiment Triplet Extraction}

\begin{abstract}
Cross-domain Aspect Sentiment Triplet Extraction (ASTE) aims to extract fine-grained sentiment elements from target domain sentences by leveraging the knowledge acquired from the source domain.
Due to the absence of labeled data in the target domain, recent studies tend to rely on pre-trained language models to generate large amounts of synthetic data for training purposes.
However, these approaches entail additional computational costs associated with the generation process.
Different from them, we discover a striking resemblance between table-filling methods in ASTE and two-stage Object Detection (OD) in computer vision, which inspires us to revisit the cross-domain ASTE task and approach it from an OD standpoint.
This allows the model to benefit from the OD extraction paradigm and region-level alignment.
Building upon this premise, we propose a novel method named \textbf{T}able-\textbf{F}illing via \textbf{M}ean \textbf{T}eacher (TFMT).
Specifically, the table-filling methods encode the sentence into a 2D table to detect word relations, while TFMT treats the table as a feature map and utilizes a region consistency to enhance the quality of those generated pseudo labels.
Additionally, considering the existence of the domain gap, a cross-domain consistency based on Maximum Mean Discrepancy is designed to alleviate domain shift problems.
Our method achieves state-of-the-art performance with minimal parameters and computational costs, making it a strong baseline for cross-domain ASTE.

\end{abstract}



\maketitle
\section{Introduction}\label{sec:introduction}
\begin{figure}[t]
    \centering
    \includegraphics[width=0.49\textwidth]{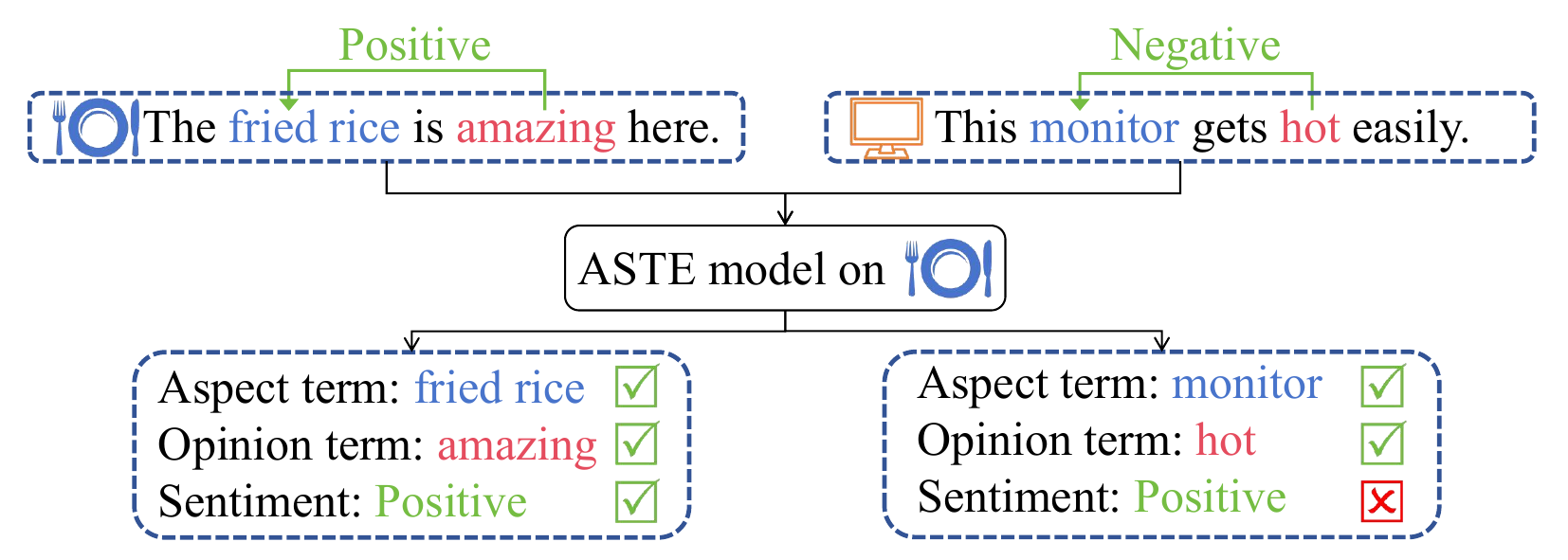}
    \vspace{-0.3cm}
    \caption{The ASTE model trained in the restaurant domain fails to produce accurate results in the laptop domain.}
    \label{fig:0}
\end{figure}

Fine-grained sentiment analysis is an important research direction in Natural Language Processing (NLP). 
In recent years, there has been a constant stream of novel tasks being presented in this field, such as aspect-based sentiment analysis (ABSA) \cite{pontiki-etal-2014-semeval,zhao2024rdgcn}, aspect opinion pair extraction (AOPE) \cite{wu-etal-2020-grid}, and aspect sentiment triplet extraction (ASTE) \cite{wu-etal-2020-grid,chen-etal-2022-enhanced,zhang-etal-2022-boundary, peng2024prompt}. 
These tasks vary in the elements they seek to extract, with ASTE representing the most intricate and challenging undertaking.
As shown in the left half of Figure~\ref{fig:0}, ASTE involves the comprehensive extraction of triplets from a given sentence, encompassing aspect terms, opinion terms, and sentiment polarity (i.e., positive, neutral, and negative). 
Give the sentence ``The \textit{fried rice} is \textit{amazing} here.'', the corresponding triplet is (\textit{fried rice}, \textit{amazing}, \textit{Positive}).
Although there have been many domain-specific ASTE works, they fail to meet the need to handle data from multiple domains in real-world scenarios.
As shown in the right half of Figure~\ref{fig:0}, when we apply the ASTE model trained exclusively in the restaurant domain to predict data from the unseen laptop domain, the model generates incorrect predictions due to its inability to adapt to the new domain knowledge.

\begin{figure*}[t]
    \centering
    \includegraphics[width=0.99\textwidth]{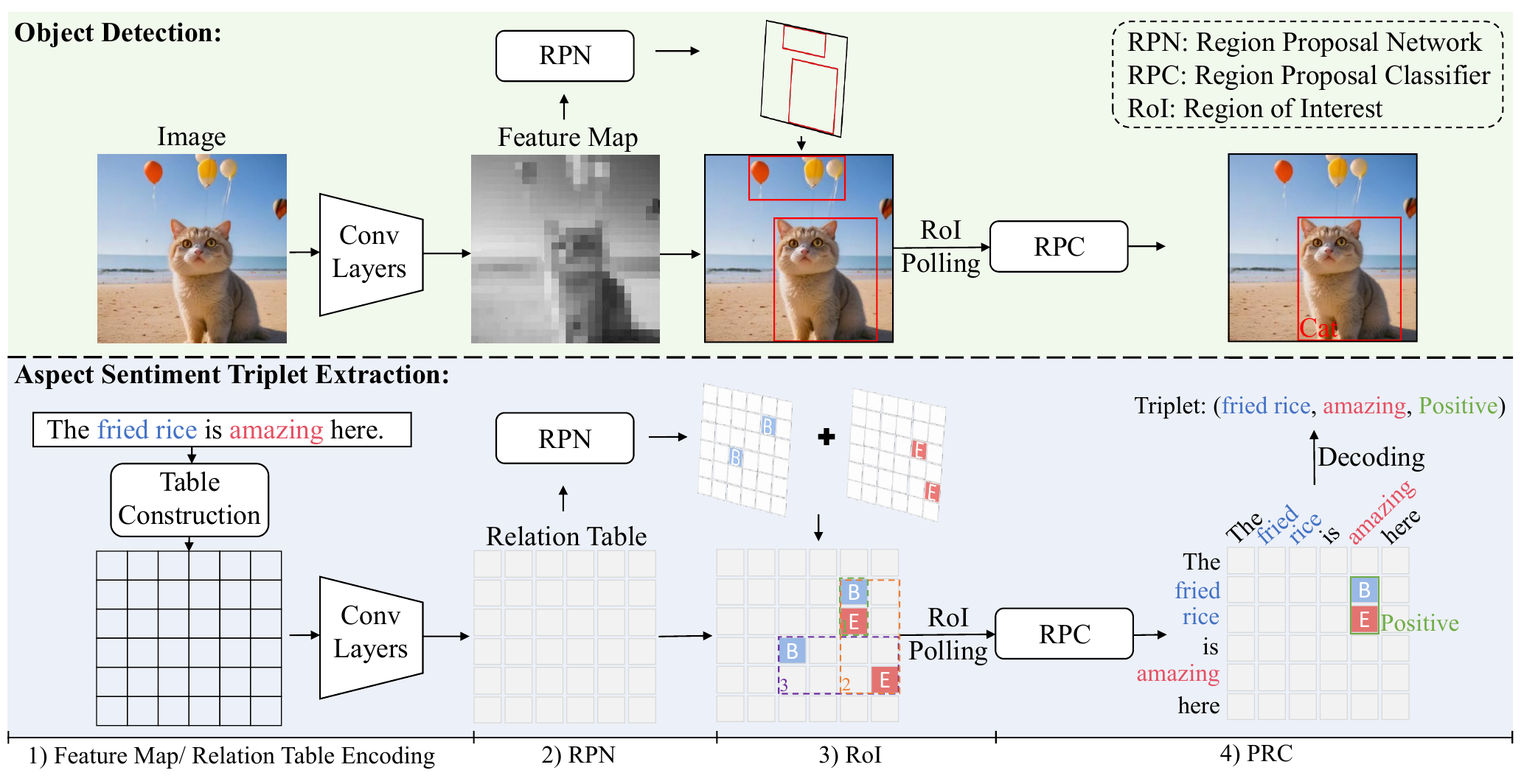}
    \caption{The general architecture of two-stage Object Detection (OD) and region-level table-filling method for Aspect Sentiment Triplet Extraction (ASTE) share similar underlying principles. $B$ (Beginning) and $E$ (Ending) represent the top-left and bottom-right corners of a candidate region, respectively.}
    \label{fig:2}
\end{figure*}

Cross-domain ASTE task \cite{deng-etal-2023-bidirectional} has newly been proposed with the objective of training models to perform ASTE tasks in the target domain, using only annotated data from the source domain and unlabeled data from the target domain.
Previous works on cross-domain fine-grained sentiment analysis focused on learning cross-domain invariant pivotal features \cite{li-etal-2019-transferable,9415156}. 
However, this approach relies on task-specific design and may not fully capture the nuanced distinctions within each particular domain.
Another more promising research direction is to leverage the power of pre-trained language models (PLMs) to generate a large amount of pseudo data that fits the distribution of the target domain \cite{li-etal-2022-generative,deng-etal-2023-bidirectional}.
Although effective, compared to providing pseudo labels for unlabeled data, this method requires significant computational costs during the data generation process.
Mean-teacher \cite{tarvainen2017mean, Cai_2019_CVPR} is a widely adopted method in unsupervised domain adaptation (UDA).
It improves the performance by training a supervised student model and an unsupervised teacher model, utilizing their prediction differences on labeled and unlabeled target domain data. 
Although highly effective in computer vision (CV) tasks, mean-teacher is rarely utilized in ASTE, where generation paradigms are generally preferred.
To leverage the embedded prior knowledge of PLMs, these methods \cite{li-etal-2022-generative,deng-etal-2023-bidirectional} choose to endure the high computational costs.
This consensus hinders the application of the mean teacher in cross-domain research within the ASTE area.
However, diverging from previous studies, we have discovered a striking similarity between the table-filling method used for ASTE and the two-stage Object Detection (OD) method \cite{DBLP:conf/nips/RenHGS15} in terms of their workflow. 
Mapping sentiment triplets to regions on the figure allows the model to benefit from staged extraction (for ASTE) and region-level alignment (for cross-domain).
This intriguing finding has inspired us to propose the \textbf{T}able-\textbf{F}illing via \textbf{M}ean \textbf{T}eacher (TFMT) approach.

Employing the table-filling method \cite{gupta-etal-2016-table,wu-etal-2020-grid} is an appealing direction for addressing the ASTE task.
It encodes a sequence of sentences into a 2D table, where the rows and columns represent different entity categories (i.e., aspect terms and opinion terms). 
The relations between entities are marked in the off-diagonal grid of the table, and supervised training is used to learn the relations between words.
%
As shown in Figure \ref{fig:2}, we observe that the region-level table-filling method shares a similar process logic with the two-stage OD, which involves four steps: 
1) A given image is encoded into a feature map using a convolutional neural network (CNN). 
2) A Region Proposal Network (RPN) is used to obtain all candidate regions. 
3) the feature vectors within each candidate region are encoded into fixed-length representations using Region of Interest (RoI) pooling. 
4) a Region Proposal Classifier (RPC) is used for region classification.
In summary, the table-filling methods treat the target triplets as specific regions within the table. To detect these regions, they first identify candidate regions and then categorize them, mirroring the two-stage process of detecting and classifying objects in images.
%
%
Many table-filling methods \cite{wu-etal-2020-grid,chen-etal-2022-enhanced,zhang-etal-2022-boundary} adopt the two-stage formulation.
Other work \cite{ning2023od} employs a one-stage extraction framework in the relation extraction task.
However, these works merely skim the surface and fail to thoroughly investigate the correlation between the two tasks or explore this direction further.
We believe that addressing cross-domain ASTE tasks from a two-stage OD perspective will yield two key advantages for the model: 1) The staged extraction paradigm will better leverage the word relations within the regions; 2) assisting cross-domain models in focusing on crucial information within the regions.

Specifically, the proposed TFMT approach utilizes a mean-teacher architecture to guide domain adaptation.
Mean-Teacher consists of two models with the same structures: a student model and a teacher model. 
The teacher model is pre-trained on the labeled source domain and then freezes its parameters in the training step, while the student model is randomly initialized.
We reformat the ASTE task into an OD task by treating the encoded table as a feature map.
The student network is supervisedly trained in each training epoch using source domain data. 
At the same time, it receives region-level consistency constraints from the teacher model's pseudo labels generated using unlabeled target domain data. 
Because the triplets are labeled as regions, region-level consistency converts the extraction alignment from student and teacher into region alignment.
After each training epoch, the teacher model parameters are updated via an exponential moving average (EMA) of the student model.
Furthermore, since the model only possesses prior knowledge from the source domain, the pseudo-labels generated on the target domain inevitably contain noise.
To alleviate this label shift phenomenon, we propose a multi-perspective consistency module based on Maximum Mean Discrepancy (MMD) \cite{pmlr-v37-long15}.

Our main contributions can be summarised as follows:

1) We observe the similarity between table-filling methods in ASTE tasks and two-stage OD, which guides us to explore an effective solution for cross-domain ASTE. 

2) The proposed multi-perspective MMD consistency can effectively mitigate the label shift caused by domain adaptation without increasing parameters.

3) The proposed TFMT framework avoids the paradigm of synthetic data generation, resulting in significantly fewer parameters and computational time costs compared to previous approaches.

4) Our framework is concise yet achieves state-of-the-art performance, serving as a strong baseline for follow-up research and inspiring the adoption of more advancing mean teacher methods in this field.


\section{Related Work}\label{sec:relatedwork}
\subsection{Aspect Sentiment Triplet Extraction (ASTE)}
ASTE is a recently emerged and challenging task.
\cite{DBLP:conf/aaai/PengXBHLS20} introduced this task and employed a pipeline approach to extract aspect terms, opinion terms, and sentiment polarities in multiple stages. 
Recognizing the heavy reliance on word-level interactions in previous studies, SpanASTE \cite{xu-etal-2021-learning} proposed a span-level method to consider span-to-span interactions explicitly.
%
Some studies have treated ASTE as a sequence-to-sequence task and utilized generative models to address it. 
GAS \cite{zhang-etal-2021-towards-generative} designed two normalized generative paradigms, annotation-style and extraction-style modeling. 
COM-MRC \cite{zhai-etal-2022-com} formalized the ASTE task into a Machine Reading Comprehension (MRC) based framework and devised a context-enhancement strategy to identify sentiment triplets.
RoBMRC \cite{liu-etal-2022-robustly} proposed a robustly optimized bidirectional machine reading comprehension method to tackle issues of query conflict and probability unilateral decrease encountered during generation.
MvP \cite{gou-etal-2023-mvp} used human-like problem-solving processes to generate and summarize sentiment elements in different orders.

Another popular approach is the table-filling \cite{miwa-sasaki-2014-modeling} method. 
GTS \cite{wu-etal-2020-grid} was the first to apply the table-filling method to the ASTE task, they annotated sentiment triplets into a two-dimensional table for extraction.
EMC-GCN \cite{chen-etal-2022-enhanced} introduced a multi-channel GCN-based approach that further integrates syntactic and textual structural information into table-filling. 
On the other hand, BDTF \cite{zhang-etal-2022-boundary} modified the original table-filling annotation scheme and proposed a boundary detection-based method.

\vspace{-0.5cm}
\subsection{UDA for Fine-grained Sentiment Analysis}
Early UDA studies for fine-grained sentiment analysis primarily focused on learning domain-invariant features. The majority of these explorations introduced syntactic knowledge as a bridge to overcome domain gaps. 
\cite{li-etal-2019-transferable} proposed a Selective Adversarial Learning (SAL) method to automatically capture the latent relations of words, thereby reducing reliance on external linguistic resources.
\cite{wang-pan-2018-recursive} developed a novel RNN that can effectively mitigate word-level domain shifts through syntactic relations.
%
EATN \cite{9415156} designed two special tasks and a novel aspect-oriented multi-head attention mechanism to learn shared features from a well-labeled source domain and guide the classification performance in the target domain.
Another line of research has placed emphasis on data.
CDRG \cite{yu-etal-2021-cross} and GCDDA \cite{li-etal-2022-generative} proposed a two-stage generative cross-domain data augmentation framework. Based on reviews labeled in the source domain, this framework first masks source-specific attributes and then converts the domain-independent review into a target-domain review, thereby generating target-domain reviews with fine-grained annotations.
BGCA \cite{deng-etal-2023-bidirectional} further advanced this approach by proposing a bidirectional generative framework. 
This framework trains generative models in text-to-label and label-to-text directions to produce higher-quality target-domain data.


\subsection{UDA for Object Detection (OD)}
To learn domain-invariant visual features, some adversarial learning methods \cite{chen2018domain,saito2019strong} employed a domain discriminator and conducted adversarial training on the feature encoder and discriminator.
Another research line is based on data augmentation. These methods \cite{chen2020harmonizing,hsu2020progressive} employed generative models (e.g., CycleGAN \cite{zhu2017unpaired}) to produce interpolated samples between the source and target domains, thereby bridging the distribution gap between the domains.
Mean-teacher \cite{tarvainen2017mean} is a semi-supervised learning method, and recently \cite{Cai_2019_CVPR} has been exploring its potential in UDA for OD.
Further investigations \cite{deng2021unbiased,li2022cross,chen2022learning} have been conducted to improve data augmentation or employ adversarial training techniques, with the goal of enhancing the quality of pseudo labels generated by the mean-teacher model. 
Contrastive Mean Teacher (CMT) \cite{cao2023contrastive} proposed an integrated framework to leverage the identified intrinsic alignment and synergy between contrastive learning and mean-teacher self-training.
\section{Method}\label{sec:method}

\begin{figure}[t]
    \centering
    \includegraphics[width=0.49\textwidth]{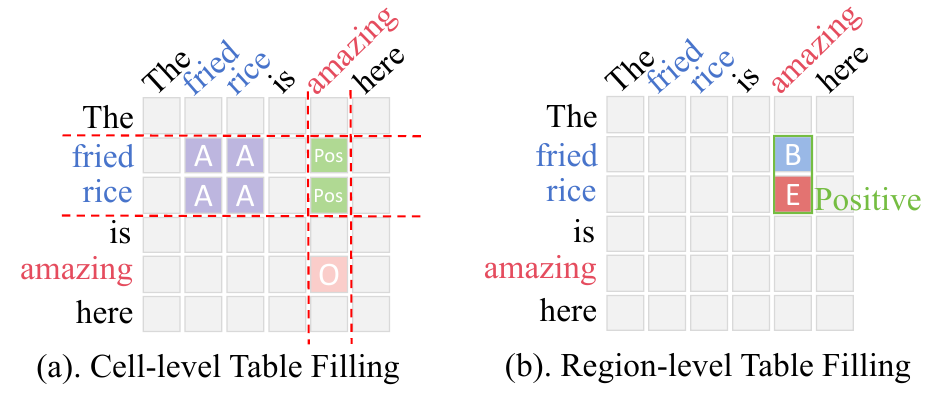}
    \vspace{-0.8cm}
    \caption{Examples of (a) cell-level and (b) region-level methods. As the latter closely resembles two-stage object detection, our TFMT is based on the region-level paradigm.}
    \label{fig:1}
\end{figure}

\begin{figure*}[t]
    \centering
    \includegraphics[width=0.95\textwidth]{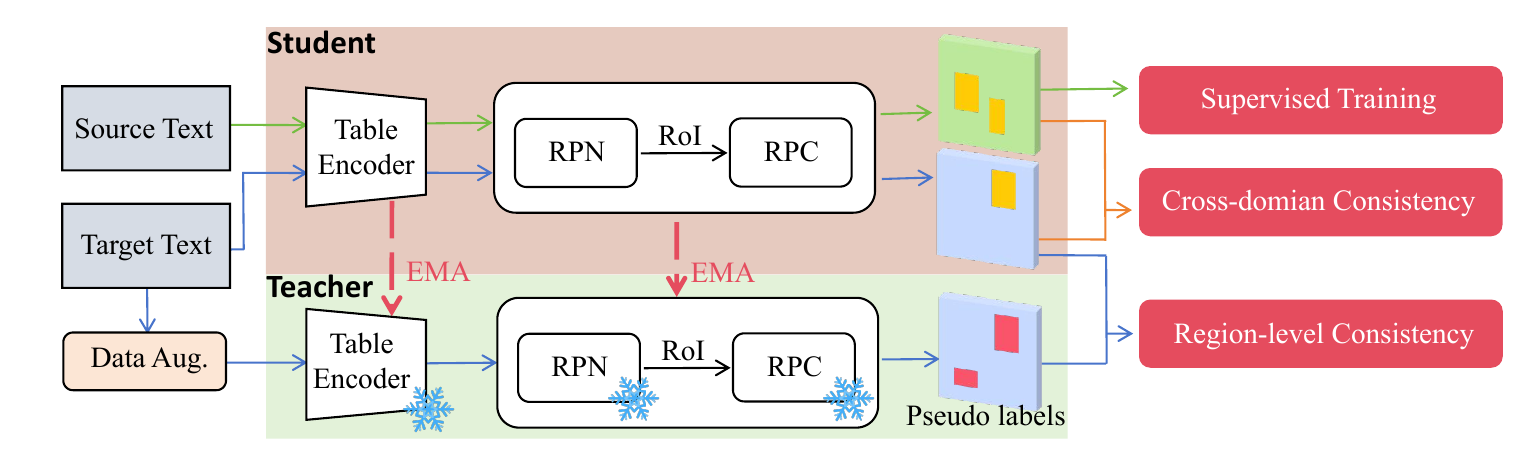}
    \vspace{-0.3cm}
    \caption{The architecture of TFMT. Our model consists of two identical architectures: 1) the teacher model, whose parameters are first pre-trained on the source domain and later frozen in each training step; 2) the student model, whose parameters are reinitialized. The student model is trained using source domain data and target domain pseudo-labels provided by the teacher model, and the teacher model parameters are updated using EMA after the training step.}
    \label{fig:3}
\end{figure*}

In this section, we first give the task formalization and introduce the table tagging scheme shown in Figure \ref{fig:1}. Then, we present the proposed Table-Filling via Mean Teacher (TFMT) method as illustrated in Figure \ref{fig:3}.

\subsection{Task Formalization}
Given a sentence $s$ with length $n$, the ASTE task aims to extract all triplets \(\tau=[(a, o, l)_1, (a, o, l)_2, ..., (a, o, l)_{|\tau|}]\) from $s$, where $a$, $o$ and $l$ denote aspect term, opinion term and sentiment polarity in $\{ Pos, Neu, Neg \}$, respectively.
For cross-domain ASTE, given a set of source domain labeled sentences $D_s={\{(s_i^s, \tau_i^s)\}_{i=1}^{|D_s|}}$ and a set of target domain unlabeled sentences $D_t={\{s_i^t\}_{i=1}^{|D_t|}}$, where $s_i^s$ and $s_i^t$ are sentences in the source and target domains, respectively. $\tau_i^s$ is the triplets set of $s_i^s$.
Our model aims to extract all triplets from the test set specific to the target domain.

\noindent \textbf{Table Tagging Scheme.}
As shown in Figure \ref{fig:1}, the relations between words in a sentence are represented in a square $n \times n$ table. 
The vertical and horizontal axes represent aspects and opinions, respectively. 
There exist two types of table-filling methods: cell-level and region-level.
As shown in Figure \ref{fig:1} (a), the cell-level method annotates aspect terms ($A$) and opinions terms ($O$) on the diagonal of the table, while the cross-region between the two is used to annotate the corresponding sentiment polarity.
It extracts the positions of aspect/opinion terms and then obtains the extraction results based on the category of the cells where they intersect. 

Unlike the cell-level method, the region-level method \cite{zhang-etal-2022-boundary} treats the target triplet as a specific region within the table and aims to precisely identify its surrounding boundaries. 
As shown in Figure \ref{fig:1} (b), regions' boundaries are labeled using beginning ($B$) and ending ($E$), followed by labeling the sentiment of the annotated regions.
Here, $B$ and $E$ are used to represent the upper left and lower right corners of a relation rectangle, respectively.
While the rest of the numerous gray cells are marked as "$None$".
This boundary-based tagging scheme converts word-level annotations in text sequences into region-level annotations on a feature map.
Unlike cell-level methods that annotate all cells at once, region-level methods aligns closely with the two-stage object detection paradigm, hence adopted in our TFMT.


\subsection{Table-Filling via Mean Teacher (TFMT)}
The architecture of TFMT\footnote{Our code and data are available at https://github.com/KunPunCN/TFMT} is shown in Figure \ref{fig:3}.
Our TFMT framework consists of a student model and a teacher model, both sharing the same model architecture. 
The student model serves as the primary model, receiving supervised training on the source domain data while also undergoing unsupervised training using unlabeled target domain data. 

The teacher model is utilized to guide the training of the student model. 
It is pretrained on the source domain and then frozen at each training step.
During training step, the teacher model focuses on the target domain and generates pseudo labels for the student model.
Specifically, in each training iteration, the target domain data $D_t$ is augmented using a random word substitution to produce $\tilde{D_t}$. 
The teacher model then processes these augmented data and produces multiple region proposals along with their predicted softmax values $\tilde{P^\tau}$.
Subsequently, we filter the region proposals and retain only those labels for which the confidence ${max}_{i\in \mathcal{C}}(\tilde{P^\tau_i})$ (where $\mathcal{C}$ is the foreground label set) exceeds the threshold $\eta$. 
This confidence threshold $\eta$ is applied to ensure that only high-confidence pseudo labels are assigned to the student model.
After each training step, The teacher's parameters are updated through an Exponential Moving Average (EMA) \cite{tarvainen2017mean} mechanism:
$\Theta_t = \lambda\Theta_t + (1-\lambda)\Theta_s$,
where $\lambda$ ($0<\lambda<1$) is a hyper-parameter. $\Theta_t$ and $\Theta_s$ denote the model parameters of teacher and student, respectively.


\subsubsection{Supervised Training}
To align with the two-stage OD paradigm for both teacher and student base models, we adopted Boundary-Driven Table-Filling (BDTF) \cite{zhang-etal-2022-boundary} and restructured the framework accordingly. 
The architecture of the base model is shown in the lower half of Figure \ref{fig:2}.


\vspace{0.2cm}
\textbf{Table Encoding.} \label{tableencoding}
Given the sentence $s=[x_1, x_2, ..., x_n]$, we input it into a pre-trained language model and get the last hidden layer $[\boldsymbol{h}_1, \boldsymbol{h}_2, ..., \boldsymbol{h}_n] \in \mathbb{R}^{n\times d} $ as the sentence representations, where $n$ is the sentence length.

We donate the relation table as $\boldsymbol{T} \in \mathbb{R}^{n\times n\times d}$, which maps the relations between any two words into separate cells.
Specifically, for a cell $\boldsymbol{t}_{ij} \in \mathbb{R}^{d}$ in $\boldsymbol{T}$, it represents the relation between the $i$-th and $j$-th words in sentence $s$.
To construct the relation table, BDTF calculates each $\boldsymbol{t}_{ij}$ as follows:
\begin{equation}
\boldsymbol{t}_{ij} = Linear(\boldsymbol{h}_i \oplus \boldsymbol{h}_j \oplus pooling(\boldsymbol{h}_{i:j}) \oplus (\boldsymbol{h}_i)^T\boldsymbol{V}\boldsymbol{h}_j),
\label{eqa}
\end{equation}
where $Linear$ denotes a nonlinear projection with an activation function, $\oplus$ denotes a concatenation operation.
The third term $pooling(\boldsymbol{h}_{i:j})$ (assume $i<j$) is a max-pooling operation, which aims to capture essential contextual information between word pairs, thereby accentuating pertinent relational features.
The fourth term $(\boldsymbol{h}_i)^T\boldsymbol{V}\boldsymbol{h}_j$ employs a tensor-based operation as described in \cite{socher2013recursive}, which is intended to harness the nuances of word-to-word interactions effectively. $\boldsymbol{V}$ is a parameter matrix.

After obtaining the relational table $\boldsymbol{T}$, we regard it as a feature map $\boldsymbol{T}^0$. To thoroughly learn and extract relational features, BDTF employs ResNet-style CNN \cite{he2016deep} as an encoder.
Specifically, given the $(l-1)$-th CNN layer representation $\boldsymbol{T}^{l-1}$, the $l$-th CNN layer representation $\boldsymbol{T}^l \in \mathbb{R}^{n\times n\times d}$ is calculated as:
\begin{equation}
\boldsymbol{T^l} = Resnet_l(\boldsymbol{T}^{l-1}),
\label{eqb}
\end{equation}
where $Resnet_l$ donates the $l$-th ResNet layer.
The design of $Resnet_l$ can be flexible, but it requires ensuring that the output dimensions of the convolutions are the same as the input dimensions $\mathbb{R}^{n\times n\times d}$.

\vspace{0.2cm}
\textbf{Region Proposal Networks (RPN).}
The RPN aims to output a set of rectangular object proposals from the table, each with an objectness score.
%
This process first requires categorizing each cell in the table as one of $\{B, E, invalid\}$. 
Given the feature map $\boldsymbol{T}^L$ output by the last convolutional layer (assume $L$ layers in total), we calculate the label probabilities $P_{ij}^B, P_{ij}^E (i,j<n)$ for each cell as:
\begin{equation}
P_{ij}^B = \sigma(Linear_B(\boldsymbol{T}^L)), \quad
P_{ij}^E = \sigma(Linear_E(\boldsymbol{T}^L)),
\end{equation}
where $Linear_B$ and $Linear_E$ are two independent fully connected layers, $\sigma$ is a sigmoid activation function.
Instead of predicting the labels with a certain threshold, BDTF uses a top-$k$ pruning strategy to retain the most promising candidates and mitigate the potential exposure bias \cite{schmidt2019generalization}. 
First, RPN sorts the cells according to $P_{ij}^B$ and $P_{ij}^E$, and then the top-$k$ cells with the highest probabilities are selected to create candidate sets $\mathcal{B}$ and $\mathcal{E}$, respectively, where $k$ is a value determined based on sentence length.

After obtaining the candidate sets $\mathcal{B}$ and $\mathcal{E}$, RPN constructs the candidate regions set based on their positional relations.
For elements $B_{(a,b)}$ (at the $(a, b)$ position) and $E_{(c,d)}$ (at the $(c, d)$ position) in $\mathcal{B}$ and $\mathcal{E}$ respectively, if $B_{(a,b)}$ is in the upper-left corner of $E_{(c,d)}$ or they overlap (i.e., $a \leq c$ and $b \leq d$), then the corresponding rectangular region $R_{abcd}$ is considered as a hit proposal.
As illustrated in the RoI stage table in Figure \ref{fig:2}, four labels $B$ and $E$ can form three corresponding candidate regions, which are respectively marked by green, orange, and purple boxes.

\vspace{0.2cm}
\textbf{RoI Polling and Region Proposal Classifier (RPC).}
For a region proposal $R_{abcd}$, we define its representation $\boldsymbol{r}_{abcd} \in \mathbb{R}^{3d}$ as:
\begin{equation}
\boldsymbol{r}_{abcd} = \boldsymbol{t}^L_{ab} \oplus \boldsymbol{t}^L_{cd} \oplus pooling(\boldsymbol{t}^L_{a:c,b:d}),
\label{eqc}
\end{equation}
where $\boldsymbol{t}^L_{ab}$ and $\boldsymbol{t}^L_{cd}$ are region vertices, $\boldsymbol{t}^L_{a:c,b:d}$ donate all the features encompassed within the region.


After obtaining the RoI representation, BTDF employs a classifier to predict its type in $\{ Pos, Neu, Neg, Invalid \}$:
\begin{equation}
P_{abcd}^\tau = softmax(Linear_P(\boldsymbol{t}_{abcd})), 
\end{equation}
where $Linear_P$ is a fully connected layer, $softmax$ is a softmax function. 
$P_{abcd}^\tau \in \mathbb{R}^{4}$ is the sentiment label of the given region.
During decoding, we drop those region proposals with $l_{abcd} = Invalid$ and generate the triplets from the remained proposals.
For instance, if region $r_{abcd}$ is predicted as $POS$, then the corresponding sentiment triplet is $(aspect = s[a:c], opinion = s[b:d], l=POS)$.



\vspace{0.2cm}
\textbf{Supervised Loss.}
Given the ground truth label $y^B_{ij}, y^E_{ij} \in \{0,1\}$ for each cell in the relation table, the loss $\mathcal{L}_{RPN}$ of RPN is calculated as the mean cross-entropy between $P_{ij}^B/P_{ij}^E$ and $y_{ij}^B/y_{ij}^E$:
\begin{equation}
\mathcal{L}_{RPN} = \sum_{i=1}^n\sum_{j=1}^nL_{CE}(P_{ij}^B, y_{ij}^B) + L_{CE}(P_{ij}^E, y_{ij}^E),
\end{equation}
where $0 \leq i,j \leq n$, $L_{CE}$ is the cross entropy loss function.
Given the ground truth label $y_i^\tau \in \{0,1,2,3\}$ for each region in the candidate proposals, the loss $\mathcal{L}_{RPC}$ of RPC is calculated as the mean cross-entropy between $P_i^\tau$ and $y_i^\tau$:
\begin{equation}
\mathcal{L}_{RPC} = -\sum_{i=1}^m L_{CE}(P_i^\tau, y_i^\tau),
\end{equation}
where $0 \leq i \leq m$. $m$ is the number of region proposals. 
The overall supervised loss $\mathcal{L}_{sup}$ is defined as the sum of $\mathcal{L}_{RPN}$ and $\mathcal{L}_{RPC}$:
\begin{equation}
\mathcal{L}_{sup} = \mathcal{L}_{RPN} + \mathcal{L}_{RPC}.
\label{eq1}
\end{equation}


\subsubsection{Region-level Consistency}
To compel alignment between teacher and student models in focusing on important information within a region, we adopt a region-level consistency as the unsupervised loss between the student and teacher models.
We define the collection of pseudo-labeled regions detected by teacher RPN as $\tilde{R}$, in both the student and teacher models, the region predictions provided by the RPN are represented as $P^\tau$ and $\tilde{P^\tau}$, respectively. 
The unsupervised loss is formulated as the Mean Squared Error between the predictions of the student and teacher:
\begin{equation}
\mathcal{L}_{uns} = \frac{1}{|\tilde{R}|}·\sum_{r \in \tilde{R}} {||P^\tau_r - \tilde{P^\tau_r}||}^2_2.
\label{eq22}
\end{equation}

Since our method labels all triplets as regions in the relation table, utilizing region-level consistency enables the alignment of predictions from student and teacher models and effectively reduces pseudo-label errors caused by unclear predicted word boundaries.

\subsubsection{Cross-domain Consistency}
Due to the fact that the knowledge learned by the teacher model mainly stems from training on the source domain labels, the use of pseudo labels generated by the teacher model in the target domain inevitably introduces noise, resulting in domain discrepancy.
Different from using a domain discriminator to align the distributions of two domains \cite{li2022cross}, we utilize the Maximum Mean Discrepancy (MMD) to minimize the domain discrepancy, which does not require any additional model parameters.
Specifically, given a source domain sample $s^s$, the student model generates a feature map $\boldsymbol{T}^s$ according to Eq. \ref{eqa} to Eq. \ref{eqb}, from which we can extract the boundary features $\boldsymbol{t}^s_{ab}$, $\boldsymbol{t}^s_{cd}$, and the region feature $\boldsymbol{r}^s_{abcd}$ for the predicted regions by Eq. \ref{eqc}.
Similarly, provided with a target domain sample $s^t$, we can extract the corresponding boundary features $\boldsymbol{t}^t_{ab}$, $\boldsymbol{t}^t_{cd}$ and the region feature $\boldsymbol{r}^t_{abcd}$.
Then we define a boundary-level MMD and a region-level MMD:
\begin{equation}
\mathcal{L}_{boundary} = L_{MMD}(\boldsymbol{t}^s_{ab}, \boldsymbol{t}^t_{ab}) + L_{MMD}(\boldsymbol{t}^s_{cd}, \boldsymbol{t}^t_{cd}),
\end{equation}
\begin{equation}
\mathcal{L}_{region} = L_{MMD}(\boldsymbol{r}^s_{abcd}, \boldsymbol{r}^t_{abcd}),
\end{equation}
where $L_{MMD}$ is the MMD function \cite{pmlr-v37-long15}.
The cross-domain consistency $\mathcal{L}_{mmd}$ is defined as the sum of the two terms:
\begin{equation}
\mathcal{L}_{mmd} = \mathcal{L}_{boundary} + \mathcal{L}_{region}.
\label{eq3}
\end{equation}

\subsubsection{Overall Loss}
The overall loss of our TFMT is:
\begin{equation}
\mathcal{L} = \mathcal{L}_{sup} + \alpha\mathcal{L}_{uns} + \beta\mathcal{L}_{mmd},
\label{eq4}
\end{equation}
where $\alpha$ and $\beta$ are the trade-off parameters, the loss terms are respectively defined in Eq. \ref{eq1}, Eq. \ref{eq22} and Eq. \ref{eq3}.
\section{Experiments}

\subsection{Experimental Settings}
\subsubsection{Datasets}
Following \cite{deng-etal-2023-bidirectional}, we evaluate our model by employing a publicly available dataset sourced from the SemEval ABSA Challenges \cite{pontiki-etal-2014-semeval,pontiki-etal-2015-semeval,pontiki-etal-2016-semeval}. 
This dataset was released by \cite{xu-etal-2020-position} and consists of four subsets, namely R14, R15, R16, and L14, which are derived from two domains: restaurant reviews and laptop reviews.
Table \ref{tab:statistics} summarizes the statistics of benchmark datasets.

\begin{table}[t]
\caption{Statistics of datasets.}
\label{tab:statistics}
\centering
\tabcolsep=5pt
\scalebox{1}{
\begin{tabular}{l|llllllll}
\hline
\multicolumn{1}{c|}{\multirow{3}{*}{Dataset}} & \multicolumn{6}{c|}{Restaurant}                                                                                                                & \multicolumn{2}{c}{laptop}                    \\
\multicolumn{1}{c|}{}                         & \multicolumn{2}{c}{R14}                       & \multicolumn{2}{c}{R15}                       & \multicolumn{2}{c|}{R16}                       & \multicolumn{2}{c}{L14}                       \\ \cline{2-9} 
\multicolumn{1}{c|}{}                         & \multicolumn{1}{c}{\#s} & \multicolumn{1}{c}{\#t} & \multicolumn{1}{c}{\#s} & \multicolumn{1}{c}{\#t} & \multicolumn{1}{c}{\#s} & \multicolumn{1}{c|}{\#t} & \multicolumn{1}{c}{\#s} & \multicolumn{1}{c}{\#t} \\ \hline
train           & 1266    & 2338       & 605  & 1013    & 857 & 1394     & 906   & 1460    \\
dev                                           & 310                   & 577                             & 148                   & 249                    & 210                   & 339  & 219                   & 346    \\
test                                          & 492                   & 994                              & 322                   & 485                    & 326                   & 514        & 328                   & 543                   \\ \hline
\end{tabular}
}
\end{table}

\begin{table*}[ht]
\centering
\caption{Results on cross-domain ASTE. The symbol $\dagger$ indicates that the results are obtained from \cite{deng-etal-2023-bidirectional}.}
\tabcolsep=5pt
\scalebox{1}{
\begin{tabular}{l|cccc|cccc|c}
\hline
Methods            & R14→L14 & R15→L14 & R16→L14 & $Avg.$ & L14→R14 & L14→R15 & L14→R16  & $Avg.$ & Total $Avg.$  \\ \hline
SpanASTE$^\dagger$ & 45.83   & 42.50   & 40.57  & 43.00  & 57.24   & 49.02   & 55.77  & 54.01  & 48.49 \\
RoBMRC$^\dagger$   & 43.90   & 40.19   & 37.81  & 40.63  & 57.13   & 45.62   & 52.05  & 51.60  & 46.12 \\
COM-MRC   & 44.87   & 40.55   & 38.23  & 41.22  & 57.42   & 47.68   & 53.11  & 52.74  & 46.98 \\
MvP      & 48.42   & 41.19   & 43.78  & 44.46  & 63.55   & 54.30   & 61.63  & 59.83  & 52.15 \\
GAS$^\dagger$      & 49.57   & 43.78   & 45.24  & 46.20  & 64.40   & 56.26   & 63.14  & 61.27  & 53.73 \\
{BGCA}$_{text \rightarrow label}^\dagger$ & 52.55   & 45.85   & 46.86  & 48.42  & 61.52   & 55.43   & 61.15  & 59.37  & 53.89 \\
{BGCA}$_{label \rightarrow text}^\dagger$ & 53.64   & 45.69   & 47.28  & 48.87  & 65.27   & 58.95   & 64.00  & 62.74  & 55.80 \\ \hline
GTS-BERT    & 43.10\scriptsize{$\pm 0.45$}   & 39.58\scriptsize{$\pm 0.30$}   & 38.34\scriptsize{$\pm 0.36$}  & 40.34\scriptsize{$\pm 0.37$}  &  50.67\scriptsize{$\pm 0.59$}  & 43.14\scriptsize{$\pm 0.37$}  & 49.47\scriptsize{$\pm 0.57$}  & 47.76\scriptsize{$\pm 0.51$} & 44.05\scriptsize{$\pm 0.44$} \\
\quad + cTFMT      & 47.92\scriptsize{$\pm 0.55$}   & 43.86\scriptsize{$\pm 0.53$}   & 42.14\scriptsize{$\pm 0.45$}  & 44.64\scriptsize{$\pm 0.51$}   & 54.86\scriptsize{$\pm 0.57$}   & 47.24\scriptsize{$\pm 0.60$}  & 52.09\scriptsize{$\pm 0.93$}  & 51.40\scriptsize{$\pm 0.70$}  & 48.02\scriptsize{$\pm 0.61$} \\
EMC-GCN      & 44.23\scriptsize{$\pm 0.48$}   & 41.64\scriptsize{$\pm 0.34$}   & 39.61\scriptsize{$\pm 0.29$}  & 41.83\scriptsize{$\pm 0.37$}   & 53.63\scriptsize{$\pm 0.51$}   & 45.09\scriptsize{$\pm 0.43$}  & 51.10\scriptsize{$\pm 0.53$}  & 49.94\scriptsize{$\pm 0.49$}  & 45.88\scriptsize{$\pm 0.43$}\\
\quad + cTFMT       & 49.79\scriptsize{$\pm 0.70$}   & 43.30\scriptsize{$\pm 0.73$}   & 42.53\scriptsize{$\pm 0.52$}  & 45.21\scriptsize{$\pm 0.65$}   & 56.82\scriptsize{$\pm 0.49$}   & 51.74\scriptsize{$\pm 0.36$}  & 54.79\scriptsize{$\pm 0.41$}  & 54.45\scriptsize{$\pm 0.42$} & 49.83\scriptsize{$\pm 0.54$}\\ \hline
BDTF         & 45.63\scriptsize{$\pm 0.67$}   & 42.37\scriptsize{$\pm 0.44$}   & 40.59\scriptsize{$\pm 0.63$}  & 42.86\scriptsize{$\pm 0.58$}   & 56.25\scriptsize{$\pm 0.76$}   & 49.57\scriptsize{$\pm 0.63$}  & 54.08\scriptsize{$\pm 0.77$}  & 53.50\scriptsize{$\pm 0.72$} & 48.08\scriptsize{$\pm 0.65$}     \\
\quad + TFMT       & \textbf{54.46}\scriptsize{$\pm 0.78$}   & \textbf{48.17}\scriptsize{$\pm 0.52$}    & \textbf{49.35}\scriptsize{$\pm 0.68$}  & \textbf{50.66}\scriptsize{$\pm 0.67$}  & \textbf{65.91}\scriptsize{$\pm 0.42$}   & \textbf{59.40}\scriptsize{$\pm 0.56$}   & \textbf{64.66}\scriptsize{$\pm 0.55$}   & \textbf{63.32}\scriptsize{$\pm 0.51$}   & \textbf{56.99}\scriptsize{$\pm 0.59$} \\ 
\hline
$\quad \Delta (Abs.\%)$  & $\uparrow$\textbf{0.82}\scriptsize{$1.53\%$}   & $\uparrow$\textbf{2.32}\scriptsize{$3.97\%$}    & $\uparrow$\textbf{2.07}\scriptsize{$4.38\%$}  & $\uparrow$\textbf{1.79}\scriptsize{$3.66\%$}  & $\uparrow$\textbf{0.64}\scriptsize{$0.98\%$}   & $\uparrow$\textbf{0.45}\scriptsize{$0.76\%$}   & $\uparrow$\textbf{0.66}\scriptsize{$1.03\%$}   & $\uparrow$\textbf{0.58}\scriptsize{$0.92\%$}    & $\uparrow$\textbf{1.16}\scriptsize{$2.08\%$}  \\ \hline
\end{tabular}
}
\label{tab:mainresult}
\end{table*}

\begin{table*}[ht]
\centering
\caption{Results on cross-domain AOPE. The symbol $\dagger$ indicates that the results are obtained from \cite{deng-etal-2023-bidirectional}.}
\tabcolsep=5pt
\scalebox{1}{
\begin{tabular}{l|cccc|cccc|c}
\hline
Methods            & R14→L14 & R15→L14 & R16→L14 & $Avg.$ & L14→R14 & L14→R15 & L14→R16 & $Avg.$ & Total $Avg.$  \\ \hline
SpanASTE$^\dagger$ & 51.90   & 48.15   & 47.30 & 49.12  & 61.97   & 55.58   & 63.26 & 60.27  & 54.69 \\
RoBMRC$^\dagger$   & 52.36   & 46.44   & 43.61 & 47.47  & 54.70   & 48.68   & 55.97 & 53.12  & 50.29 \\
COM-MRC   & 53.36   & 45.79   & 44.14  & 47.76  & 54.58   & 49.27   & 56.63  & 53.49  & 50.63 \\
GAS$^\dagger$      & 57.58   & 53.23   & 52.17 & 54.33  & 64.60   & 60.26   & 66.69 & 63.85  & 59.09 \\
GCDDA & 59.17   & 54.83   & 53.19 & 55.73  & 64.42   & 62.11   & 68.05 & 64.86  & 60.30 \\
{BGCA}$_{text\rightarrow label}^\dagger$ & 58.54   & 54.06   & 51.99 & 54.86  & 64.61   & 58.74   & 67.19 & 63.51  & 59.19 \\
{BGCA}$_{label\rightarrow text}^\dagger$ & 60.82   & 55.22   & 54.48 & 56.84  & 68.04   & 65.31   & 70.34 & 67.80  & 62.37 \\ \hline
GTS-BERT         & 48.14\scriptsize{$\pm 0.58$}   & 44.87\scriptsize{$\pm 0.43$}   & 43.27\scriptsize{$\pm 0.55$}  & 45.43\scriptsize{$\pm 0.52$}   & 56.52\scriptsize{$\pm 0.50$}   & 49.44\scriptsize{$\pm 0.53$}  & 57.54\scriptsize{$\pm 0.68$}  & 54.50\scriptsize{$\pm 0.57$} & 49.96\scriptsize{$\pm 0.55$}\\
\quad + cTFMT           & 55.39\scriptsize{$\pm 0.45$}   & 52.59\scriptsize{$\pm 0.48$}   & 52.94\scriptsize{$\pm 0.66$}  & 53.64\scriptsize{$\pm 0.53$}   & 63.31\scriptsize{$\pm 0.44$}   & 56.75\scriptsize{$\pm 0.51$}  & 64.92\scriptsize{$\pm 0.43$}  & 61.66\scriptsize{$\pm 0.46$} & 57.65\scriptsize{$\pm 0.50$}\\
EMC-GCN          & 50.43\scriptsize{$\pm 0.39$}   & 43.62\scriptsize{$\pm 0.29$}   & 43.87\scriptsize{$\pm 0.31$}  & 45.97\scriptsize{$\pm 0.33$}   & 53.04\scriptsize{$\pm 0.49$}   & 49.45\scriptsize{$\pm 0.38$}  & 58.74\scriptsize{$\pm 0.36$}  & 53.74\scriptsize{$\pm 0.41$} & 49.86\scriptsize{$\pm 0.37$}\\
\quad + cTFMT           & 55.81\scriptsize{$\pm 0.58$}   & 52.41\scriptsize{$\pm 0.70$}   & 52.16\scriptsize{$\pm 0.55$}  & 53.46\scriptsize{$\pm 0.61$}   & 64.06\scriptsize{$\pm 0.35$}   & 58.12\scriptsize{$\pm 0.44$}  & 65.31\scriptsize{$\pm 0.41$}  & 62.50\scriptsize{$\pm 0.40$} & 57.98\scriptsize{$\pm 0.51$}\\ \hline
BDTF             & 50.77\scriptsize{$\pm 0.48$}   & 47.31\scriptsize{$\pm 0.50$}   & 45.38\scriptsize{$\pm 0.61$}  & 47.82\scriptsize{$\pm 0.53$}   & 61.29\scriptsize{$\pm 0.31$}   & 53.92\scriptsize{$\pm 0.24$}  & 61.83\scriptsize{$\pm 0.41$}  & 59.01\scriptsize{$\pm 0.32$} & 53.42\scriptsize{$\pm 0.43$} \\
\quad + TFMT           & \textbf{62.50}\scriptsize{$\pm 0.53$} & \textbf{55.87}\scriptsize{$\pm 0.75$}  & \textbf{56.58}\scriptsize{$\pm 0.73$} & \textbf{58.32}\scriptsize{$\pm 0.67$} & \textbf{69.10}\scriptsize{$\pm 0.69$} & \textbf{66.67}\scriptsize{$\pm 0.61$} & \textbf{70.62}\scriptsize{$\pm 0.50$} & \textbf{68.80}\scriptsize{$\pm 0.60$} & \textbf{63.56}\scriptsize{$\pm 0.64$} \\
\hline
$\quad \Delta (Abs.\%)$  & $\uparrow$\textbf{1.68}\scriptsize{$2.76\%$}   & $\uparrow$\textbf{0.65}\scriptsize{$1.18\%$}    & $\uparrow$\textbf{2.10}\scriptsize{$2.60\%$}  & $\uparrow$\textbf{1.48}\scriptsize{$1.56\%$}  & $\uparrow$\textbf{1.06}\scriptsize{$1.56\%$}   & $\uparrow$\textbf{1.36}\scriptsize{$2.08\%$}   & $\uparrow$\textbf{0.28}\scriptsize{$0.40\%$}   & $\uparrow$\textbf{0.90}\scriptsize{$1.33\%$}    & $\uparrow$\textbf{1.19}\scriptsize{$1.91\%$}  \\ \hline
\end{tabular}
}
\label{tab:mainresult2}
\end{table*}

\subsubsection{Implementation}
We opt for BERT \cite{vaswani2017attention} as our language encoder and a two-layer ResNet \cite{he2016deep} as the feature encoder, with a dimension set at 768.
The pruning threshold $k$ and the confidence threshold $\eta$ are set to 0.3 and 0.98, respectively.
We directly use BERT-base-uncased \footnote{https://huggingface.co/bert-base-uncased} as the pretraining parameters to initialize both the student and teacher models. 
If not specified, the default setting for the rate smoothing coefficient $\lambda$ of the EMA is 0.6.
The loss weight factors $\alpha=1$ and $\beta=0.005$.
For the data augmentation method, since the prerequisite for region-level consistency is to maintain the same encoded length of the augmented sentences, we randomly replaced 50\% of the words in the sentence with words that have the same encoding lengths.
We conduct model training on one RTX 3090 GPU for 10 epochs, utilizing a batch size of 4 and setting the learning rate to $3\times 10^{-5}$. 
We assess the performance of the training model on the development set and save the best one. 
We use sentence-level F1 score as the evaluation metric. In this metric, a sentence is considered a true positive only when all sentiment triplets within it are correctly extracted.
All reported results are averaged across five runs with different random seeds. 

\vspace{0.2cm}
\textbf{Cell-level TFMT (cTFMT).}
The cell-level method doesn't utilize a region proposal method like the two-stage OD paradigm but instead directly classifies each cell in the relation table.
To adapt to the TFMT method, we treat each cell as an independent region (in $\mathcal{L}_{uns}$ and $\mathcal{L}_{mmd}$) and keep the other TFMT modules (i.e., $\mathcal{L}_{sup}$) unchanged.
Specifically, in Eq. \ref{eq22}, $\tilde{R}$ represents all cells predicted as one of $\tilde{c}=\{A, O, POS, NEG, NEU\}$. 
In Eq. \ref{eq3}, $\mathcal{L}_{mmd}$ is modified to the sum of MMD for all types of cells:
\begin{equation}
\mathcal{L}_{mmd} = \sum_{k \in \tilde{c}}{L_{MMD}(\boldsymbol{t}^s_{k}, \boldsymbol{t}^t_{k})}.
\end{equation}
The cell-level TFMT (cTFMT) is a degradation of the original region-level TFMT.

\subsubsection{Baselines}
Although there are many baseline models for cross-domain sentiment analysis (e.g., EATN \cite{9415156}, CDRG \cite{yu-etal-2021-cross}, GCDDA \cite{li-etal-2022-generative} and BGCA \cite{deng-etal-2023-bidirectional}), these works mainly focus on sentence-level sentiment analysis or aspect-based sentiment analysis (ABSA). Among them, BGCA addresses both ASTE and AOPE tasks, while GCDDA specifically targets the AOPE task.

Due to the scarcity of work on UDA for ASTE currently, besides BGCA, we have also leveraged state-of-the-art models in a zero-shot manner within the in-domain setting. 
These models have been categorized into three groups: 

1) \textbf{pipeline}: SpanASTE \cite{xu-etal-2021-learning} captures the span-to-span interactions during the multi-step filtering process. 

2) \textbf{generative}: COM-MRC \cite{zhai-etal-2022-com} and RoBMRC \cite{liu-etal-2022-robustly} address the ASTE task in a Machine Reading Comprehension (MRC) manner. GAS \cite{zhang-etal-2021-towards-generative}, GCDDA\cite{li-etal-2022-generative}, MvP \cite{gou-etal-2023-mvp}, and BGCA \cite{deng-etal-2023-bidirectional} views it as an index generation task. 

3) \textbf{table-filling}: GTS-BERT \cite{wu-etal-2020-grid} and EMC-GCN \cite{chen-etal-2022-enhanced} follow the cell-level paradigm, while BDTF \cite{zhang-etal-2022-boundary} follows the region-level paradigm.
In the above methods, the pipeline and table-filling methods are based on BERT, while the generation methods are based on t5-base\footnote{https://huggingface.co/google-t5/t5-base}.

\subsection{Main Results} \label{rq1}
The main results are reported in Table \ref{tab:mainresult}, where the best results are highlighted in bold. 
According to the results, our findings can be summarized as follows:
1) Our top-performing model (BDTF-TFMT) outperforms the current SOTA baseline BGCA$_{label\rightarrow text}$ across all tasks, with an average improvement of 1.79\% on the restaurant-to-laptop dataset, 0.58\% on the laptop-to-restaurant dataset, and an overall average improvement of 1.16\%.
This demonstrates the superiority of our BDTF-TFMT model.
2) The TFMT architecture brings about significant improvements to the table-filling model, with respective enhancements of 3.97\%, 3.95\%, and 8.83\% observed for GTS-BERT, EMC-GCN, and BDTF models. 
This highlights the adaptability of the TFMT framework across different table-filling methods.
Additionally, the most notable improvements are observed in the BDTF model, which aligns closely with the two-stage OD paradigm and thereby benefits the most from the region-level consistency component integrated within TFMT.
We will further substantiate this finding through subsequent comparative experiments.


\vspace{0.2cm}
\textbf{Result on the cross-domain AOPE task.}
In order to provide comprehensive evidence of the TFMT method's enhancement in cross-domain fine-grained sentiment analysis, we conduct an extra experiment on the aspect opinion pair extraction (AOPE) task with the same dataset.
Compared to ASTE, AOPE does not require sentiment polarity extraction; instead, its aim is to extract all (aspect term, opinion term) pairs from a given sentence.
To align with the AOPE extraction paradigm, we modify the classifier in the original RPC, which determines the sentiment polarity of regions, to a binary classifier that determines the region's validity.

Table \ref{tab:mainresult2} shows the result of the cross-domain AOPE task, we have the following findings:
1) Our best-performing BDTF-TFMT model still comprehensively outperforms all the baselines. 
Compared to BGCA$_{label\rightarrow text}$, it achieves an average improvement of 1.48\% on the restaurant-to-laptop dataset and 0.90\% on the laptop-to-restaurant dataset, resulting in an overall average improvement of 1.19\%.
2) TFMT also demonstrates superior performance on BDTF (average gain of 10.14\%) compared to GTS-BERT (average gain of 7.69\%) and EMC-GCN (average gain of 8.12\%).

We observe that the findings from Table \ref{tab:mainresult2} and Table \ref{tab:mainresult} are consistent, with similar conclusions drawn from the former clearly evident in the latter.
These results demonstrate that TFMT provides benefits when tackling cross-domain fine-grained sentiment analysis tasks using OD formulas.

\subsection{Computational Costs Analysis}\label{rq2}
We conduct a comparative experiment on computational costs.
The average results are shown in Table~\ref{tab:costs}.
From  the table, we have the following findings:
1) Comparing the results of the table-filling model before and after using TFMT, as TFMT is a pluggable module designed for guiding cross-domain learning, it is only utilized during the model training process, thus having no impact on the model's inference speed. 
Furthermore, since TFMT does not introduce any additional parameters, the model's parameter count does not change.
2) Compared to BGCA$_{label\rightarrow text}$, BDTF-TFMT exhibits a 19.1\% reduction in average training time per sample due to the fact that BGCA spends more time on pseudo-data generation.
3) In comparison to BGCA$_{label\rightarrow text}$, BDTF-TFMT has 27.35\% fewer parameters, yet achieves a 2.08\% higher F1 score and the average inference time per sample decreases by 65.04\%.
This can be attributed to the fact that while BGCA operates on a generative backbone (encoder-decoder architecture, e.g., t5), the table-filling method relies on an auto-encoder backbone (encoder-only architecture, e.g., BERT).
Consequently, the table-filling method typically entails lower parameter count and inference costs.

\begin{table}[t]
\centering
\caption{Comparison of computational costs for the cross-domain ASTE task. 
\#Params. denotes the parameter count measured in millions (M). 
\#Train and \#Infer denote the time cost of training and inference, respectively. ``/ms'' denotes the average processing time per sample.
``BGCA$_{l\rightarrow t}$'' denotes {BGCA}$_{label\rightarrow text}$.
$Abs.\%$ represents the absolute gain of BDTF+TFMT compared to BGCA$_{l\rightarrow t}$.
}
\tabcolsep=2.5pt
\scalebox{1}{
\begin{tabular}{l|ccccc}
\hline
Methods  & backbone                   & $Avg.F1$ & \#Params.             & \#Train   & \#Infer                   \\ \hline
BGCA$_{l\rightarrow t}$     & t5-base                    & 55.80  & 223M                  & 81.78 /ms & 9.21 /ms                  \\ \hline
GTS-BERT  &  \multirow{6}{*}{bert-base}    & 44.05  & \multirow{2}{*}{162M} & 69.17 /ms & \multirow{2}{*}{4.41 /ms} \\
\quad + cTFMT    &             & 48.02  &         & 72.27 /ms &                           \\
EMC-GCN &  & 45.88  & \multirow{2}{*}{121M} & 49.06 /ms & \multirow{2}{*}{2.63 /ms} \\
\quad + cTFMT    &               & 49.83  &                  & 51.13 /ms &                           \\
BDTF     &                            & 53.42  & \multirow{2}{*}{162M} & 64.24 /ms & \multirow{2}{*}{3.22 /ms} \\
\quad + TFMT    &                            & 56.99  &                       & 66.16 /ms &                           \\ \hline
$Abs.\%$  & - & $\uparrow$\textbf{2.08$\%$} & $\downarrow$\textbf{27.35$\%$}    & $\downarrow$\textbf{19.10$\%$} & $\downarrow$\textbf{65.04$\%$}    \\ \hline
\end{tabular}
}
\label{tab:costs}
\end{table}

\begin{table}[t]
\centering
\caption{Comparative study.}
\tabcolsep=3.5pt
\begin{tabular}{cl|cccc|c}
\hline
\multicolumn{2}{c|}{\multirow{2}{*}{Method}} & \multicolumn{2}{c|}{ASTE}        & \multicolumn{2}{c|}{AOPE} & \multirow{2}{*}{$Avg.$} \\
\multicolumn{2}{c|}{}                        & R$\rightarrow$ L   & \multicolumn{1}{c|}{L$\rightarrow$ R} & R$\rightarrow$ L         & L$\rightarrow$ R         &                       \\ \hline
\multirow{2}{*}{GTS-BERT}     & + cTFMT       & 44.64 & 51.40   & 53.64       & 61.66       & 52.84      \\
                              & + s-t         & 45.91 & 54.71   & 54.59       & 62.15       & 54.34                 \\ \cline{1-2}
\multirow{2}{*}{EMC-GCN}      & + cTFMT       & 45.21 & 54.45   & 53.46       & 62.50       & 53.91      \\
                              & + s-t         & 46.96 & 57.35   & 52.37       & 65.13       & 55.45                 \\ \cline{1-2}
\multirow{3}{*}{BDTF}         & + TFMT       & \textbf{50.66} & \textbf{63.32}   & \textbf{58.32}       & \textbf{68.80}       & \textbf{60.28}      \\
                              & + s-t         & 48.17 & 59.65   & 54.57       & 65.91       & 57.08                 \\
                              & + cTFMT  & 43.16 & 52.23   & 51.49       & 60.65       & 51.88                \\ \hline
\end{tabular}
\label{tab:ablation}
\end{table}

\subsection{Comparative Study} \label{rq3}
To investigate the mechanism of the TFMT module, we conduct a comparative experiment, and the results are reported in Table~\ref{tab:ablation}. 
The designs for different domain adaptation components here in Table~\ref{tab:ablation} are as follows:

\begin{itemize}[leftmargin=*]
    \item \textbf{TFMT/cTFMT}: The effect exhibited by TFMT and cTFMT.
    \item \textbf{s-t}: Replace TFMT with the self-training method. Similar to the mean-teacher approach, the model is first pre-trained on the source domain in this setup. Subsequently, the model generates pseudo-labels for the target domain data and incorporates them into the training of the next iteration. Since there are no paired teacher-student models, $\mathcal{L}_{uns}$ and $\mathcal{L}_{mmd}$ cannot be implemented.
\end{itemize}
Based on the data presented in the table, we can observe that both GTS-BERT and EMC-GCN exhibit lower TFMT performance compared to the self-training method (averaging 1.5\% and 1.54\%, respectively). 
In contrast, BDTF shows a significantly higher TFMT performance than the self-training method (averaging 3.2\%). 
We speculate that this discrepancy arises from the fact that GTS-BERT and EMC-GCN do not align with the two-stage OD paradigm, resulting in the degradation of the region-level consistency component in the original TFMT.
To validate our hypothesis, we conduct additional ablation experiments on BDTF, as shown in the table under the ``BDTF + cTFMT" row.

\begin{itemize}[leftmargin=*]
    \item \textbf{BDTF + cTFMT}: In this setting, we replace the region-level architecture with a cell-level architecture that is similar to GTS (as depicted in Figure~\ref{fig:1}(a)).
Additionally, we keep the encoding component of the model unchanged.
Since we treat each cell in the table as an independent region proposal, the original region-level consistency loss degrades to a cell-level consistency loss: $\mathcal{L'}_{uns} = \frac{1}{|C|} \sum_{k \in C} {||P_{k} - \tilde{P_{k}}||}^2_2,$
where $C$ donates the set of cells with high-confidence pseudo labels.
\end{itemize}

As a consequence of degradation, we can observe that the ``cTFMT" configuration showcases a decrease in performance of 8.4\% compared to ``TFMT" and a decrease of 5.2\% compared to ``self-training''.

\begin{figure}[t]
    \centering
    \vspace{-0.2cm}
    \includegraphics[width=0.5\textwidth]{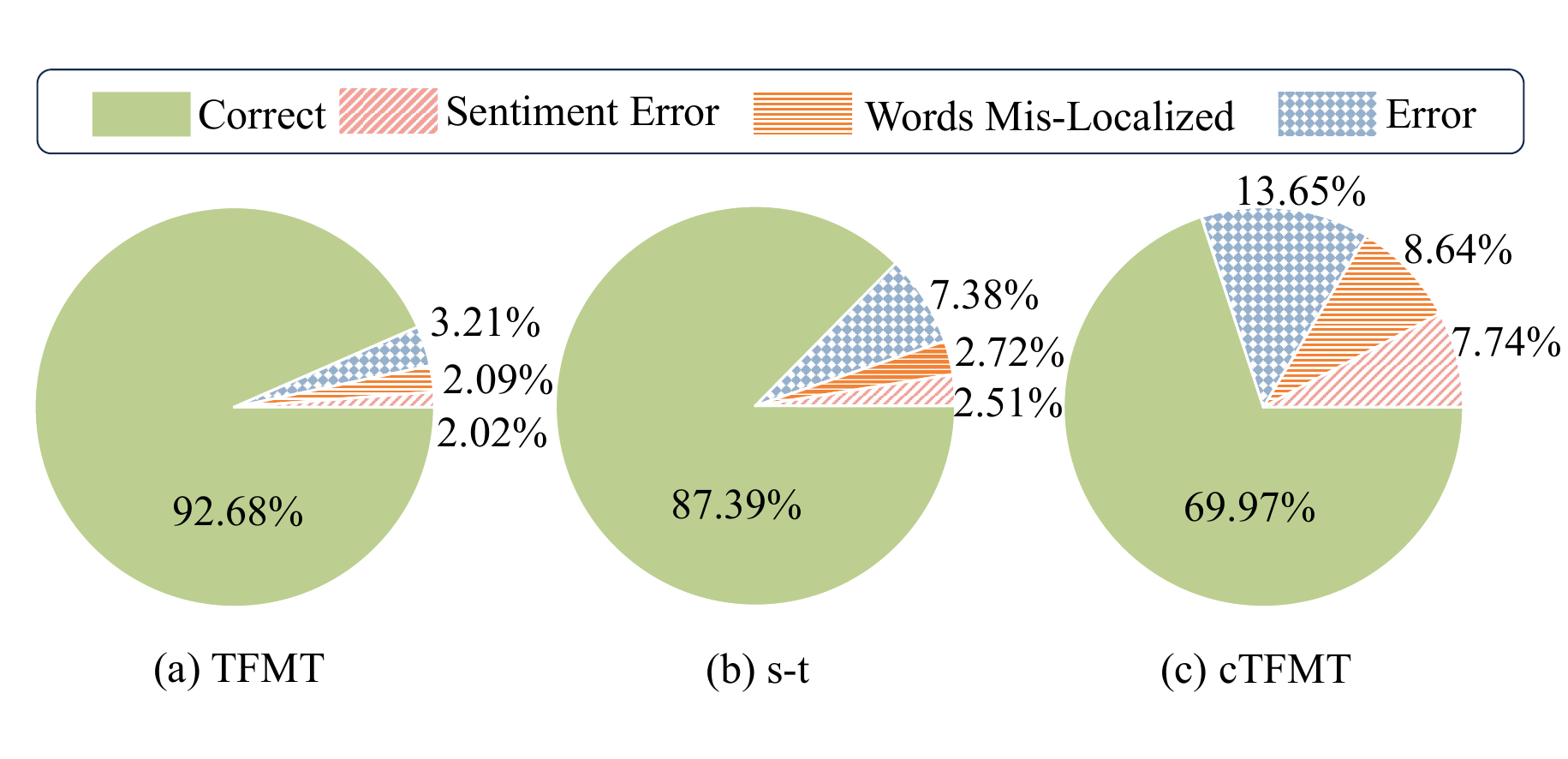}
    \vspace{-1cm}
    \caption{Error analysis of pseudo labels.}
    \label{fig:5}
\end{figure}

\subsection{Error Analysis of pseudo labels} \label{rq4}
As the quality of pseudo labels significantly influences cross-domain performance, we further investigate the effect of TFMT by examining the generation of pseudo labels in three settings: ``TFMT'', ``s-t'', and ``cTFMT''. 
The results are presented in Figure~\ref{fig:5}. We categorize all types of errors into four types: 

\begin{itemize}[leftmargin=*]
    \item \textbf{Correct}: Aspect term, opinion term, and sentiment polarity in the pseudo triplet are all accurate.

    \item \textbf{Sentiment Error}: Aspect term and opinion term are accurate, but sentiment polarity is incorrect.

    \item \textbf{Words Mis-Localized}: Sentiment polarity is correct, but the indexing for aspect term or opinion term is inaccurate.

    \item \textbf{Error}: Aspect term or opinion term is incorrect.
\end{itemize}

From the experimental results, we have the following findings:
1) The TFMT method effectively reduces the occurrence of incorrect pseudo-labels, resulting in a respective increase of 5.29\% and 22.71\% in the number of correctly labeled pseudo-labels compared to the s-t method and cTFMT method.
2) Compared to the s-t method, the TFMT method excels mainly in reducing the occurrence of error-type pseudo-labels.
This can be attributed to the incorporation of Cross-Domain Consistency $\mathcal{L}_{mmd}$ to constrain domain differences, thereby leading to a noticeable decrease in incorrectly labeled samples.
3) The cTFMT method exhibits more erroneous labels, attributed to its deviation from the object detection paradigm, where Region-level Consistency $\mathcal{L}_{uns}$ and Cross-Domain Consistency $\mathcal{L}_{mmd}$ fail to yield substantial effects.



\subsection{More Analysis} \label{rq5}
\vspace{0.2cm}
\quad \textbf{Ablation Study.}
To delve into the module mechanisms within TFMT, we conduct a series of ablation experiments as shown in Table \ref{tab:ablation2}, and we have the following observations:
1) When the data augmentation module is removed, the model's performance does not exhibit a significant decline. 
This is because the primary aim of data augmentation is to introduce perturbations to the teacher model's inputs, thereby mitigating the risk of overfitting. 
2) When we remove $\mathcal{L}_{uns}$ or $\mathcal{L}_{mmd}$ individually, there is a noticeable decline in model performance. 
Removing $\mathcal{L}_{uns}$ led to performance drops of 0.93\% and 1.49\% on the cross-domain ASTE and AOPE tasks, respectively. 
Similarly, removing $\mathcal{L}_{mmd}$ results in performance decreases of 1.65\% and 1.13\% on the respective tasks.
When $\mathcal{L}_{uns}$ and $\mathcal{L}_{mmd}$ both are removed simultaneously, the model's performance severely degrades. 
The performance decreases of 3.14\% and 3.69\% on the respective tasks.
This demonstrates the significant role played by both $\mathcal{L}_{uns}$ and $\mathcal{L}_{mmd}$ in TFMT.

\begin{table}[t]
\caption{Ablation study. ``w/o Aug.'', ``w/o $\mathcal{L}_{uns}$'' and ``w/o $\mathcal{L}_{mmd}$'' donate removing the data augmentation module, $\mathcal{L}_{uns}$ and $\mathcal{L}_{mmd}$, respectively.}
\begin{tabular}{lccccl}
\hline
                           & Task & R$\rightarrow$ L & L$\rightarrow$ R & $Avg.F1$ & $\Delta$ \\ \hline
\multirow{2}{*}{BDTF-TFMT} & ASTE & 50.66        & 63.32   &56.99  & - \\
                           & AOPE & 58.32    & 68.80       & 63.56   & - \\ \cline{2-6} 
\multirow{2}{*}{-w/o Aug.}        & ASTE & 50.27    & 62.47   & 56.37   & 0.62  \\
                           & AOPE & 58.10    & 67.82   & 62.96   & 0.60  \\
\multirow{2}{*}{-w/o $\mathcal{L}_{uns}$}        & ASTE & 49.51  & 62.61  & 56.06   &  0.93 \\
                           & AOPE & 57.34    & 66.80   & 62.07   &  1.49 \\
\multirow{2}{*}{-w/o $\mathcal{L}_{mmd}$}        & ASTE & 49.17  & 61.89  & 55.53   & 1.65  \\
                           & AOPE & 57.85    & 67.01    & 62.43  &  1.13 \\ 
\multirow{2}{*}{-w/o $\mathcal{L}_{uns}$ \& $\mathcal{L}_{mmd}$}        & ASTE & 48.67  & 59.02    & 53.85  &  3.14 \\
                           & AOPE & 53.69      & 66.05        & 59.87  & 3.69  \\ \hline
\end{tabular}
\label{tab:ablation2}
\end{table}

\begin{figure}[t]
    \centering
    \includegraphics[width=0.5\textwidth]{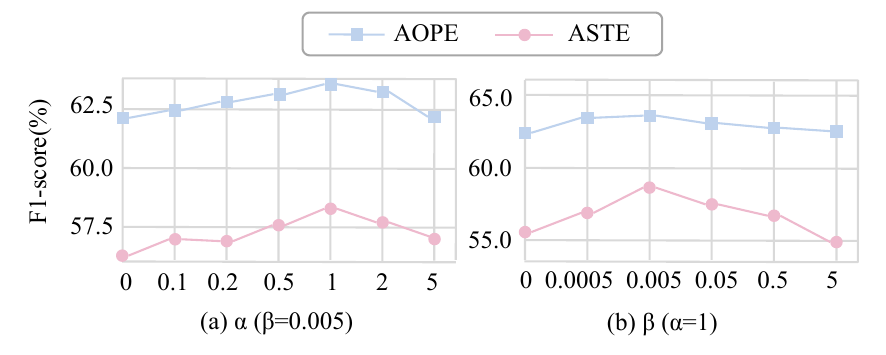}
    \vspace{-0.6cm}
    \caption{Effects of $\alpha$ and $\beta$.}
    \label{fig4}
\end{figure}

\vspace{0.2cm}
\textbf{Coefficients Study.}
To investigate the effect of weight coefficients $\alpha$ and $\beta$ in Equation \ref{eq4}, we present the average performance curves in Figure~\ref{fig4}.
In Figure~\ref{fig4}(a), fixing $\beta$ at 0.005, we find the optimal performance when $\alpha=1$.
In Figure~\ref{fig4}(b), setting $\alpha$ as 1, we observe that the best performance is attained when $\beta = 0.005$.
Moreover, excessively large values of $\beta$ can introduce an ambiguous decision boundary problem \cite{saito2018maximum} and cause a decrease in the model's classification performance.

\section{Conclusion}
In this paper, we find the inherent consistency between table-filling methods for ASTE and two-stage OD methods. 
This insight motivates our proposed TFMT method to perceive the ASTE task as an OD task and then leverages region-level consistency constraints for unsupervised training.
Experimental results show that the proposed method significantly improves boundary-based paradigms, while the grid-based paradigm leads to a degradation in region-level consistency, further underscoring the alignment of the two tasks.
To improve the quality of pseudo labels, we propose an MMD-based consistency constraint, which effectively mitigates the issue of domain discrepancy. 
Our method achieves state-of-the-art performance while operating under significantly reduced parameter counts and computational costs compared to previous approaches. 
This makes it an ideal and strong baseline for subsequent work in this field.
In future work, we will continue to develop further advanced mean-teacher methods for cross-domain ASTE.


\section*{Acknowledgments}
This work was supported by Pilot Projects of Chinese Academy of Sciences (No.E3C0011101), NSFC through grant 62322202, Shijiazhuang Science and Technology Plan Project through grant 231130459A, and Guangdong Basic and Applied Basic Research Foundation through grant 2023B1515120020.

\newpage



\bibliographystyle{ACM-Reference-Format}
\bibliography{sample-base}


\end{document}